\documentclass{article}

\usepackage{microtype}
\usepackage{graphicx}
\usepackage{subfigure}
\usepackage{booktabs}
\usepackage{multirow}
\usepackage{multicol}
\usepackage{makecell}
\usepackage{times}
\usepackage{latexsym}
\usepackage{amsmath}
\usepackage{amsfonts}
\usepackage{algorithm}
\usepackage{algpseudocode}
\usepackage{siunitx}
\usepackage{booktabs} 
\usepackage{caption}
\usepackage{hyperref}

\usepackage[accepted]{mlsys2025}

\mlsystitlerunning{Fast Forward}

\begin{document}

\twocolumn[
\mlsystitle{Fast Forward: Accelerating LLM Prefill with Predictive FFN Sparsity}



\mlsyssetsymbol{equal}{*}

\begin{mlsysauthorlist}
\mlsysauthor{Aayush Gautam}{tam,intern}
\mlsysauthor{Mukul Gagrani}{qual}
\mlsysauthor{Junyoung Park}{qual}
\mlsysauthor{Mingu Lee}{qual}
\mlsysauthor{Chris Lott}{qual}
\mlsysauthor{Narasimha Reddy}{tam}
\end{mlsysauthorlist}

\mlsysaffiliation{tam}{Department of Electrical and Computer Engineering, Texas A \& M Univeristy, Texas, USA}
\mlsysaffiliation{qual}{Qualcomm Technologies, Inc.}
\mlsysaffiliation{intern}{Author completed the research that
is the basis of this paper during an internship at Qualcomm Technologies, Inc.}

\mlsyscorrespondingauthor{Aayush Gautam}{aayushgautam@tamu.edu}

\mlsyskeywords{Machine Learning, MLSys}

\vskip 0.3in

\begin{abstract}
The prefill stage of large language model (LLM) inference is a key computational bottleneck for long-context workloads. At short-to-moderate context lengths (1K–16K tokens), Feed-Forward Networks (FFNs) dominate this cost, accounting for most of the total FLOPs. Existing FFN sparsification methods, designed for autoregressive decoding, fail to exploit the prefill stage’s parallelism and often degrade accuracy. To address this, we introduce FastForward, a predictive sparsity framework that accelerates LLM prefill through block-wise, context-aware FFN sparsity. FastForward combines (1) a lightweight expert predictor to select high-importance neurons per block, (2) an error compensation network to correct sparsity-induced errors, and (3) a layer-wise sparsity scheduler to allocate compute based on token-mixing importance. Across LLaMA and Qwen models up to 8B parameters, FastForward delivers up to 1.45× compute-bound speedup at 50\% FFN sparsity with $<$ 6\% accuracy loss compared to the dense baseline on LongBench, substantially reducing Time-to-First-Token (TTFT) for efficient, long-context LLM inference on constrained hardware.
\end{abstract}
]



\printAffiliationsAndNotice{}  
\raggedbottom
\section{Introduction}

Large Language Models (LLMs) are increasingly deployed across a wide range of real-world applications such as summarization, coding assistance, email composition, transcription, and information retrieval. While larger models consistently achieve superior performance across benchmarks, their substantial computational and memory requirements pose significant challenges for inference and deployment, especially on resource-constrained hardware like phones and laptops. Even smaller variants within the same model family—though offering competitive accuracy—face deployment bottlenecks due to limited compute throughput and memory bandwidth.

Transformers~\citep{vaswani2023attentionneed} have become the foundational architecture of modern LLMs. The original transformer introduced an encoder–decoder structure, but subsequent decoder-only variants, popularized by GPT models~\citep{radford_narasimhan_salimans_sutskever_2018}, now dominate open-weight LLM designs. Each forward pass through a decoder-only transformer involves a stack of $L$ transformer blocks, each composed of two major sublayers: \textbf{(i)} a self-attention layer that performs \emph{token mixing}, and \textbf{(ii)} a feed-forward network (FFN) that performs \emph{channel mixing}. Both sublayers are dominated by large matrix multiplications, yet their computational characteristics differ substantially.

LLM inference proceeds in two distinct stages: \textit{prefill} and \textit{decoding}. The prefill stage encodes the input prompt in a single, highly parallel forward pass. This stage is compute-bound, and its latency—often referred to as the \textit{Time-To-First-Token (TTFT)}—directly affects user experience in interactive applications such as chatbots and code copilots. In contrast, the decoding stage generates tokens autoregressively, one at a time. Since each decoding step reuses model parameters and KV cache with limited arithmetic intensity, it is typically memory-bandwidth bound. Despite these differences, both stages derive their computational load primarily from the transformer’s attention and FFN layers.

To reduce TTFT, it is crucial to identify the dominant source of computation within the prefill stage. The self-attention mechanism incurs a cost dominated by the query–key product $QK^\top$, which scales quadratically with sequence length $T$, yielding complexity $O(T^2 d_{\text{head}}\times h) \approx O(T^2 d_{\text{model}}) $. In contrast, the FFN applies a non-linear transformation that scales linearly with sequence length, $O(T d_{\text{model}} d_{\text{ffn}})$. Although linear in $T$, the FFN remains the most expensive component at short and moderate sequence lengths because its intermediate dimension $d_{\text{ffn}}$ is substantially larger—typically 4$\times$ or $\frac{8}{3}\times$—the model’s hidden size $d_{\text{model}}$.

\begin{figure}
    \centering
    \includegraphics[width=\linewidth]{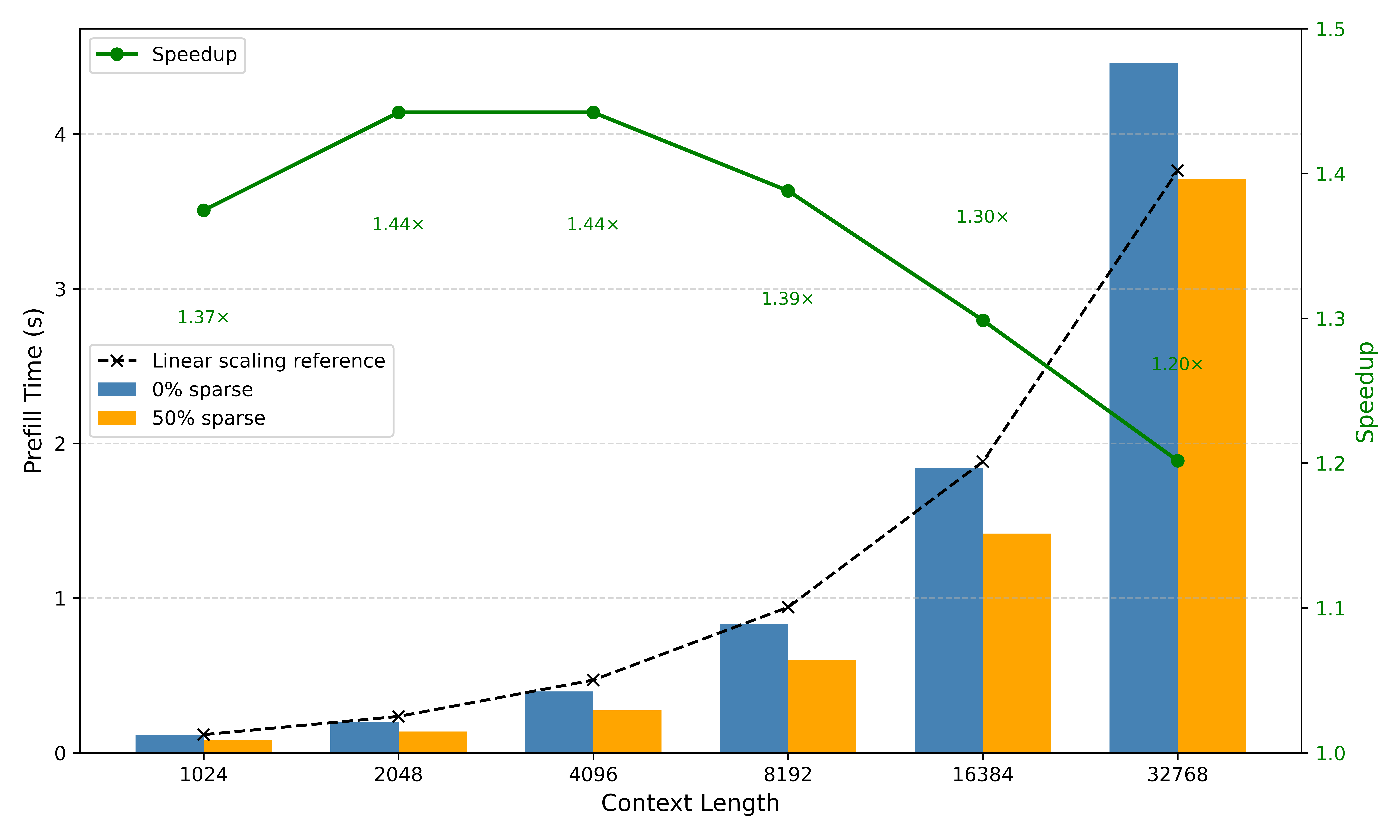}
    \caption{Time to First Token (TTFT) for the LLaMA-3.1-8B-Instruct model with and without 50\% sparsity across different context lengths, measured on a single NVIDIA A100 GPU.}
    \vspace{-3mm}
    \label{fig:prefill_times}
\end{figure}

To make this trade-off concrete, consider Llama-3.1-8B, which has $d_{\text{model}}=4096$ and $d_{\text{ffn}}=14336$. For this configuration, FFN operations dominate overall FLOPs until the sequence length exceeds approximately 28,000 tokens. Consequently, FFN layers can account for more than two-thirds of the total computation and parameters during inference. This observation implies that for short-to-moderate context lengths—typically up to 28K tokens in real-world applications—the FFN’s large constant-factor compute cost outweighs the quadratic cost of attention. Figure~\ref{fig:prefill_times} illustrates this behavior: TTFT scales approximately linearly with sequence length at short and moderate contexts, where FFN computation dominates. Thus, optimizing FFN computation is the most impactful strategy for accelerating the prefill stage. In this work, we propose to tackle this bottleneck by introducing structured sparsity into the FFN layers, significantly reducing prefill latency while preserving model accuracy.


\begin{table}[h]
\centering
\small
\setlength{\tabcolsep}{2.8pt} 
\begin{tabular}{@{}lccc@{}}
\toprule
\textbf{Workload} &
\makecell{\textbf{Prompt}\\\textbf{Len}} &
\makecell{\textbf{Output}\\\textbf{Len}} &
\makecell{\textbf{Prompt:}\\\textbf{Decode}} \\
\midrule
\makecell[l]{Programming\\(\citeauthor{hendrycks2021measuringcodingchallengecompetence},~\citeyear{hendrycks2021measuringcodingchallengecompetence})} &
3871 $\pm$ 1656 &
190 $\pm$ 343 &
20.4:1 \\

\makecell[l]{Tool Use\\(\citeauthor{guo2025stabletoolbenchstablelargescalebenchmarking},~\citeyear{guo2025stabletoolbenchstablelargescalebenchmarking})} &
1835 $\pm$ 742 &
43 $\pm$ 16 &
42.7:1 \\

\makecell[l]{Embodied Agent\\(\citeauthor{shridhar2021alfworldaligningtextembodied},~\citeyear{shridhar2021alfworldaligningtextembodied})} &
2285 $\pm$ 471 &
16 $\pm$ 13 &
142.8:1 \\
\bottomrule
\end{tabular}
\caption{Prompt and output lengths (in tokens) for representative LLM workloads, adapted from \citet{srivatsa2024prebleefficientdistributedprompt}. The large prompt-to-decode ratios highlight the compute intensity of the prefill stage in realistic LLM applications.}
\label{tab:prompt_decode_ratios}
\end{table}

While extensive research has targeted the memory-bound decoding stage through techniques such as parallel decoding~\citep{leviathan2023fastinferencetransformersspeculative, delcorro2023skipdecodeautoregressiveskipdecoding}, contextual sparsity~\citep{liu2023dejavucontextualsparsity, dong2024promptpromptedadaptivestructuredpruning, lee2024catscontextuallyawarethresholdingsparsity, akhauri2024shadowllmpredictorbasedcontextualsparsity, zhang2025rsparserankawareactivationsparsity, sparseinfer, garcia2025adaptiverankallocationspeeding}, and selective attention~\citep{leviathan2025selectiveattentionimprovestransformer, zhang2023h2oheavyhitteroracleefficient, tang2024questqueryawaresparsityefficient}, the compute-bound prefill stage remains a critical performance bottleneck—particularly for long-context tasks and deployment on edge devices. This criticality of the prefill computation is further reflected in real-world workload distributions. Empirical analyses of production-scale LLM deployments show that prompts are often significantly longer than generated outputs. \citet{qiao2025swiftkvfastprefilloptimizedinference} report that enterprise workloads such as retrieval-augmented generation (RAG) and code completion exhibit average prompt-to-generation ratios of roughly 10:1. In some domains, this ratio can reach as high as 142:1 ~\citep{srivatsa2024prebleefficientdistributedprompt}. As summarized in Table \ref{tab:prompt_decode_ratios}, this imbalance underscores the practical importance of optimizing the prefill stage. Moreover, the majority of prompts in these workloads fall between 2K and 4K tokens—precisely the range where our proposed sparsity-based method achieves the greatest acceleration (Figure~\ref{fig:prefill_times}).

Prior efforts to reduce Time to First Token (TTFT) have primarily focused on complementary directions such as prompt compression \cite{jiang2023llmlinguacompressingpromptsaccelerated, liu2025speculativeprefillturbochargingttft}, attention optimization \cite{chen2021scatterbrainunifyingsparselowrank, yang2024tidaldecodefastaccuratellm}, and layer skipping \cite{jaiswal2024ffnskipllmhiddengemautoregressive, He_Yao_Zuo_Gao_Li_Zheng_Wu_2025, luo-etal-2025-diffskip}. Another line of work leverages static sparsity through weight pruning \cite{frantar2023sparsegptmassivelanguagemodels, yin2025outlierweighedlayerwisesparsity} and matrix decompositions \cite{ashkboos2024slicegptcompresslargelanguage}, though these approaches generally underperform compared to contextual sparsity methods \cite{liu2023dejavucontextualsparsity}. However, the Feed-Forward Network (FFN)—one of the most compute-intensive components during prefill—has received limited attention. Existing FFN sparsification methods—developed largely for the autoregressive decoding phase—struggle to translate effectively to prefill. For example, CATS \cite{lee2024catscontextuallyawarethresholdingsparsity} applies token-level weight masking, which breaks the inherent parallelism advantage of prefill, while GRIFFIN \cite{dong2024promptpromptedadaptivestructuredpruning} depends on prompt-level statistics to estimate neuron importance—information unavailable prior to prompt processing. Moreover, some transformer layers perform more work than others and uniformly applied sparsity across long prompts often leads to cumulative error and degraded accuracy.

In contrast, Fast Forward introduces a prefill-specific framework that directly addresses these shortcomings. Instead of relying on post-hoc neuron selection or static sparsity, it predicts neuron importance proactively through a lightweight expert predictor, enabling block-wise adaptive sparsification that preserves parallelism. Furthermore, an error compensation network mitigates sparsity-induced degradation, while a layer-wise sparsity schedule dynamically allocates computational budgets to more influential layers. Together, these components enable efficient FFN sparsity tailored to prefill, achieving substantial TTFT reduction without sacrificing model fidelity.

In this paper, we propose a novel framework for FFN sparsification tailored for deployment on devices with limited memory and compute resources. Our approach includes:
\begin{itemize}
    \item \textbf{Block-wise Prompt Processing:} Following \cite{agrawal2023sarathiefficientllminference,fast-ondevice}, we partition the prompt into 128-token blocks to mitigate the activation memory overhead, enabling efficient processing of long prompts on devices with limited memory capacity.
    \item \textbf{Expert Neuron Prediction:} For each block, we use a lightweight neural network to predict the most relevant FFN neurons, enabling targeted sparsification.
    \item \textbf{Error Compensation Network:} A small auxiliary model is trained to correct errors introduced by sparsification, preserving output fidelity.
    \item \textbf{Dynamic Sparsity Control:} We adapt sparsity levels across layers based on the attention scores received by non-sink tokens, ensuring that more critical layers receive higher computational fidelity.
\end{itemize}
This framework enables efficient LLM inference on edge devices by reducing TTFT without compromising model accuracy, making short-to-moderate context processing more practical and responsive. While our primary focus is TTFT optimization, the approach generalizes to decoding-time acceleration as well.

\section{Background: Transformer Block Computation}

A Transformer block comprises a sequence of operations—\textbf{Layer Normalization}, \textbf{Causal Self-Attention}, \textbf{Residual Connections}, and \textbf{Feed-Forward Network (FFN)} transformations.
Each component operates on a sequence of token embeddings represented as vectors of dimension $d_{\text{model}}$. Among these, the \textit{Causal Self-Attention} and \textit{Feed-Forward Network} layers dominate the overall computational cost.

\subsection{Causal Self-Attention}

Given an input sequence $X \in \mathbb{R}^{T \times d_{\text{model}}}$, where $T$ is the number of tokens, the model first projects each token representation into \textit{Query (Q)}, \textit{Key (K)}, and \textit{Value (V)} matrices:
\begin{equation}
Q = X W_Q, \quad K = X W_K, \quad V = X W_V,
\end{equation}
where $W_Q, W_K, W_V \in \mathbb{R}^{d_{\text{model}} \times d_k}$ are learnable projection matrices and $d_k$ denotes the per-head dimensionality. In practice, multi-head attention is used with $h$ parallel heads, where the total model dimension is partitioned across heads such that $h \times d_k = d_{\text{model}}$.

The projection step requires approximately
\begin{equation}
O(T \times d_{\text{model}}^2)
\end{equation}
floating-point operations (FLOPs).

The attention scores are computed as:
\begin{equation}
A = \text{softmax}\!\left(\frac{Q K^\top}{\sqrt{d_k}} + M\right),
\end{equation}
where $M$ is a causal mask ensuring that each token attends only to previous tokens.  
The matrix multiplication $Q K^\top$ requires
\begin{equation}
O(T^2 \times d_{\text{model}})
\end{equation}
operations. The attention output is then obtained as:
\begin{equation}
O = A V,
\end{equation}
which adds another
\begin{equation}
O(T^2 \times d_{\text{model}})
\end{equation}
FLOPs.  
Overall, the Causal Self-Attention layer scales quadratically with sequence length, making it one of the main computational bottlenecks in Transformers.

\subsection{Feed-Forward Network (FFN)}

The FFN applies two linear transformations separated by a non-linear activation function $\sigma(\cdot)$ (typically GELU or ReLU). For input $X \in \mathbb{R}^{T \times d_{\text{model}}}$:
\begin{equation}
H = \sigma(X W_{\text{up}}), \quad
Y = H W_{\text{down}},
\end{equation}
where
\begin{equation}
W_{\text{up}} \in \mathbb{R}^{d_{\text{model}} \times d_{\text{ffn}}}, \quad
W_{\text{down}} \in \mathbb{R}^{d_{\text{ffn}} \times d_{\text{model}}}.
\end{equation}
The FFN typically expands the hidden dimension to $d_{\text{ffn}} \approx 4 d_{\text{model}}$, resulting in a computational cost of:
\begin{equation}
O(T \times d_{\text{model}} \times d_{\text{ffn}}).
\end{equation}

In gated variants (e.g., Gated Linear Units), an additional projection $W_{\text{gate}}$ is introduced:
\begin{equation}
Y = \left[\sigma(X W_{\text{gate}}) \odot (X W_{\text{up}})\right] W_{\text{down}},
\end{equation}
which approximately doubles the cost of the first projection but preserves the same asymptotic complexity.

\subsection{Comparative Computational Cost}

The computational cost of a Transformer block can be approximately expressed as the sum of its attention and feed-forward network (FFN) components:
\begin{equation}
C_{\text{total}} = O(T^2 \times d_{\text{model}}) + O(T \times d_{\text{model}} \times d_{\text{ffn}}),
\end{equation}
where the first term captures the quadratic dependence of self-attention on the sequence length $T$, and the second term scales linearly with $T$ but quadratically with the hidden dimensions.

When $T$ is relatively small (e.g., $T < c \times d_{\text{ffn}}$, where $c \approx 2$), the FFN component dominates the overall computation. Given that $d_{\text{ffn}}$ is typically large—ranging from $8192$ in the LLaMA-3.2-1B model to $14336$ in the 8B variant—the FFN remains the primary computational bottleneck for sequence lengths up to approximately $16$K tokens for the 1B model and $28$K tokens for the 8B model. These ranges encompass most practical prompt lengths encountered during inference.

Therefore, for short to moderate context lengths, FFN computation constitutes the dominant factor affecting Time-to-First-Token (TTFT) latency. This observation motivates our focus on improving the efficiency of FFN computations to achieve faster model response times.

\begin{figure}[!htpb]
    \vspace{-3mm}
    \centering
    \includegraphics[width=0.95\linewidth]{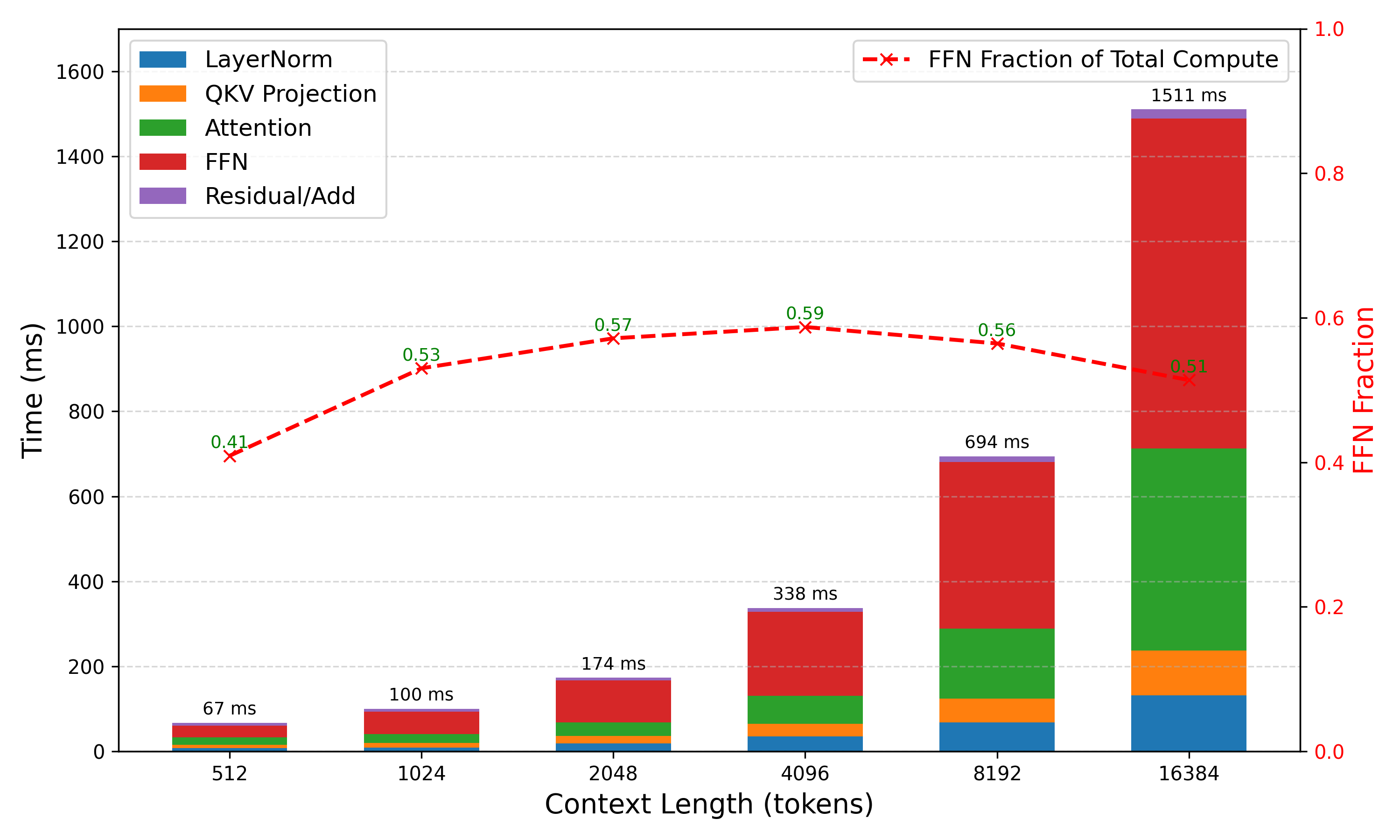}
    \caption{Time taken by different components of a transformer block to process tokens of varying context length for Llama-3.1-8B profiled in a single A100 GPU}
    \vspace{-5mm}
    \label{fig:ffn_component}
\end{figure}



\section{Methodology}

\begin{figure*}[!ht]
    \centering
    \includegraphics[width=0.65\linewidth]{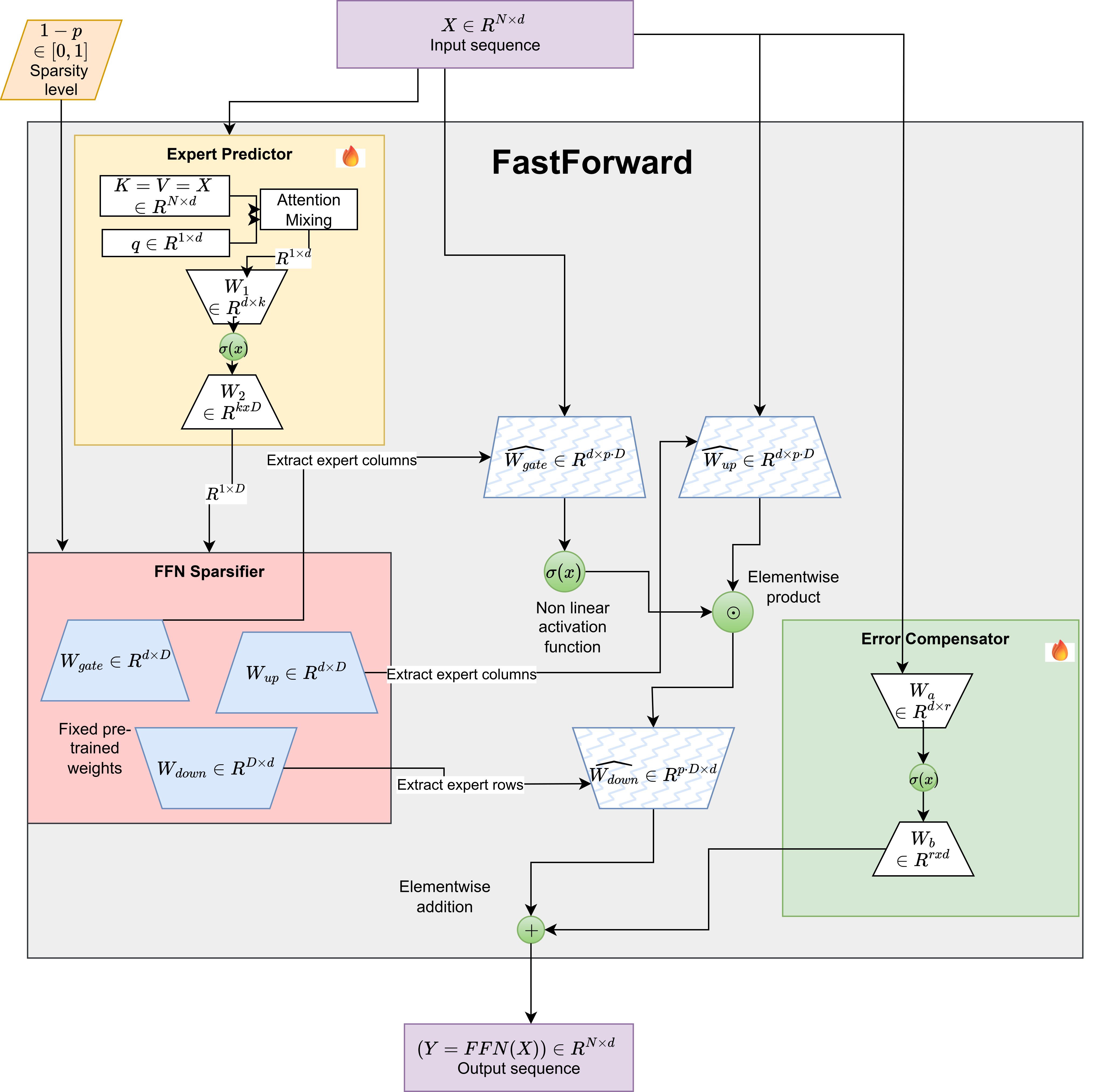}
    \caption{Block diagram showing the working of FastForward}
    \vspace{-5mm}
    \label{fig:blockdiagram-efficientFFN}
\end{figure*}

\subsection{Block-wise Prompt Processing}

Edge devices often operate under stringent memory constraints. For instance, processing a full prompt of length 16K with a model like LLaMA-3.2-3B can require over 1GB of activation memory. To mitigate this, prompts are typically processed block-by-block on NPUs, as demonstrated in works such as \cite{park2025keydiffkeysimilaritybasedkv, chen2021scatterbrainunifyingsparselowrank, fast-ondevice}. This approach significantly reduces activation memory requirements.
A block size of 128 tokens is commonly used \cite{fast-ondevice}, striking a balance between keeping NPU compute cores fully utilized and maintaining memory overhead within a few megabytes.

This block-wise processing paradigm opens up opportunities for sparsity during the prefill phase. GRIFFIN \cite{dong2024promptpromptedadaptivestructuredpruning} observed a phenomenon termed \textit{flocking}, where certain intermediate neurons in the MLP layers exhibit consistently high activations across all prompt tokens. These neurons act as "experts" for the given prompt context. By activating only a subset (e.g., 50\%) of these expert neurons during generation, GRIFFIN achieves minimal accuracy loss on classification and summarization benchmarks.

Our research extends this concept to the prefill stage by sparsely computing the FFN projections in the MLP layers during prompt processing. Unlike GRIFFIN, which uses a static set of experts derived from prompt statistics for all decoding steps, our approach adaptively selects the top-$K$ expert neurons for each block, allowing finer-grained specialization during both prefill and decoding. This introduces several challenges and opportunities:

\begin{enumerate}
    \item \textbf{Expert Prediction:} Since block-specific experts are not known a priori, we train an expert predictor—a lightweight attention-based module that takes the same input as the FFN and predicts the most important neurons for that block.
    
    \item \textbf{Error Compensation:} As context length increases and deeper layers are reached, sparsification errors accumulate, leading to non-negligible accuracy degradation. To address this, we introduce a low-rank error compensator module that estimates the residual error from the FFN inputs and offsets the sparse FFN outputs accordingly.
    
    \item \textbf{Layerwise Sparsity Scheduling:} We observe that the first block often contains \textit{sink tokens}, which are crucial for maintaining stable attention score distributions across varying context lengths \cite{xiao2024efficientstreaminglanguagemodels, barbero2025llmsattendtoken}. Additionally, attention scores across layers are unevenly distributed among non-sink tokens, indicating varying layer importance in token embedding transformations. This motivates a layerwise sparsity schedule to optimize performance at fixed sparsity levels.
\end{enumerate}

\subsection{Expert Neuron Predictor}
\label{sec:expert_pred}

We propose a lightweight \textit{expert neuron predictor} that identifies the most relevant neurons for each block of tokens, enabling block-wise sparse FFN computation during prompt processing. Unlike prior approaches that use a fixed expert set throughout decoding, our predictor dynamically selects neurons at each block, adapting to input semantics while maintaining low overhead.

\paragraph{Architecture.}
The expert predictor consists of two key components:
\vspace{-2mm}
\begin{itemize}
    \item \textbf{Attention module:} A single-head attention layer with a trainable query vector \( q_{\text{pred}} \in \mathbb{R}^{1 \times d_{\text{model}}} \). Given the block input \( X^{l,i} \in \mathbb{R}^{N_{\text{block}} \times d_{\text{model}}} \), token embeddings serve as both keys and values, and the attention output is computed as:
    \begin{equation}
        a = \text{Softmax}\!\left( \frac{q_{\text{pred}} X^{l,i^\top}}{\sqrt{d_{\text{model}}}} \right) X^{l,i} \in \mathbb{R}^{1 \times d_{\text{model}}}
    \end{equation}
    where \( a \) represents an aggregated block representation.
    \item \textbf{Feedforward module:} A two-layer MLP that projects \( a \) into the FFN neuron space:
    \begin{equation}
        s = \text{ReLU}\!\left(a W_{\text{pred}_1}\right) W_{\text{pred}_2}
    \end{equation}
    where \( W_{\text{pred}_1} \in \mathbb{R}^{d_{\text{model}} \times r} \) and \( W_{\text{pred}_2} \in \mathbb{R}^{r \times d_{\text{ffn}}} \), with the reduced dimension \( r = d_{\text{model}}/16 \) (rounded up to the nearest power of two, i.e., \(128\) for LLaMA-1B and \(256\) for LLaMA-3B/8B).
\end{itemize}
\vspace{-2mm}
\paragraph{Expert selection.}
The output vector \( s \in \mathbb{R}^{1 \times d_{\text{ffn}}} \) provides an activation score for each FFN neuron. We obtain the binary expert mask \( M \in \{0,1\}^{d_{\text{ffn}}} \) as:
\begin{equation}
    M_j = 
    \begin{cases}
        1, & \text{if } s_j \text{ is among the top-} K \text{ scores}\\
        0, & \text{otherwise}
    \end{cases}
\end{equation}

Given \( M \), we select the corresponding rows and columns from the FFN weight matrices:
\begin{align}
\hat{W}_{\text{up}} &= \text{SelectRows}(W_{\text{up}}, M) \in \mathbb{R}^{K \times d_{\text{model}}} \\
\hat{W}_{\text{gate}} &= \text{SelectRows}(W_{\text{gate}}, M) \in \mathbb{R}^{K \times d_{\text{model}}} \\
\hat{W}_{\text{down}} &= \text{SelectCols}(W_{\text{down}}, M) \in \mathbb{R}^{d_{\text{model}} \times K}
\end{align}

The sparsified FFN computation for block \( i \) at layer \( l \) then becomes:
\begin{equation}
\begin{split}
\widehat{\text{FFN}}(X^{l,i}) = \hat{W}_{\text{down}} \cdot 
\Big( & \sigma(\hat{W}_{\text{gate}} X^{l,i}) \odot \\
& (\hat{W}_{\text{up}} X^{l,i}) \Big)
\end{split}
\end{equation}

\paragraph{Training.}
The expert predictor is trained using a weighted binary cross-entropy (BCE) objective, where labels are derived from neuron activation magnitudes following the approach in GRIFFIN. Specifically, activations with the highest norms in the dense FFN are labeled as positive (1), and the rest as negative (0):
\begin{itemize}
    \item The top 50\% of activations are labeled 1, and the bottom 50\% as 0.
    \item Among positives, weights decay exponentially: the top 20\% receive a weight of 32, the next 20\% a weight of 16, and so on.
    \item All negatives are assigned a weight of 1.
\end{itemize}

The overall loss for layer \( l \) is thus:
\begin{equation}
    \mathcal{L}_{\text{pred}}^{(l)} = - \sum_{j=1}^{d_{\text{ffn}}} w_j \left[ y_j \log \sigma(s_j) + (1 - y_j) \log (1 - \sigma(s_j)) \right]
\end{equation}
where \( y_j \in \{0,1\} \) is the binary label, \( w_j \) the sample weight, and \( \sigma \) the sigmoid function.

This training encourages the predictor to assign high scores to neurons with strong activations while maintaining differentiable learning dynamics. Once trained, the predictor operates independently during inference, incurring minimal overhead while enabling efficient block-wise sparse FFN computation.
\vspace{-2mm}

\subsection{Error Compensation Network}
\label{sec:error_comp}

Sparsifying the FFN introduces projection errors that can accumulate across layers and degrade model accuracy. To mitigate these effects, we introduce a lightweight \textit{error compensation network} that learns to predict and correct the residual error introduced by sparsification. The network operates in parallel with the sparsified FFN, producing a corrective signal that restores the fidelity of the block representation with minimal computational overhead.

\paragraph{Architecture.}
The compensator is a two-layer feedforward network with an intermediate dimension \( r' = d_{\text{model}}/8 \), chosen to balance expressiveness and efficiency. Given the block input \( X^{l,i} \in \mathbb{R}^{N_{\text{block}} \times d_{\text{model}}} \), the network output is formulated as:
\begin{equation}
Y_{\text{comp}} = W_{\text{comp}_2} \cdot \sigma \left( W_{\text{comp}_1} \cdot X^{l,i} \right)
\end{equation}
where \( W_{\text{comp}_1} \in \mathbb{R}^{d_{\text{model}} \times r'} \) and \( W_{\text{comp}_2} \in \mathbb{R}^{r' \times d_{\text{model}}} \) are trainable projection matrices.  
The final corrected output of the sparsified FFN becomes:
\begin{equation}
Y = \widehat{\text{FFN}}(X^{l,i}) + Y_{\text{comp}}
\end{equation}

\paragraph{Training.}
We train the error compensation network using \textit{layerwise distillation}. For each transformer layer, the sparse FFN output is compared against the corresponding dense FFN output from the teacher model. The compensator is optimized to minimize the mean squared error (MSE) between the dense activation \( Y_{\text{dense}} \) and the compensated sparse activation \( Y \):
\begin{equation}
\mathcal{L}_{\text{comp}}^{(l)} = \| Y_{\text{dense}} - Y \|_2^2
\end{equation}

To stabilize training, we employ a two-phase schedule. The compensator is first \textit{warm-started} using oracle expert masks—where the true top-$K$ neurons (by activation magnitude) are known—before transitioning to masks predicted by the learned expert predictor. This allows the compensator to learn meaningful error patterns prior to adapting to the predictor’s sparsification behavior.

Training is performed on the Minipile dataset for 10{,}000 steps with a batch size of 512.

\vspace{-2mm}
\subsection{Dynamic Sparsity Control}

We observe that the first block of tokens frequently contains \textit{sink tokens}, which are essential for maintaining stable attention score distributions across varying context lengths~\cite{xiao2024efficientstreaminglanguagemodels, barbero2025llmsattendtoken}. Additionally, attention scores across layers are unevenly distributed among non-sink tokens, suggesting that different layers contribute unequally to token embedding transformations. This motivates the use of a layerwise sparsity schedule to optimize performance under fixed sparsity constraints.

\begin{figure}[h]
    \centering
    \includegraphics[width=0.95\linewidth]{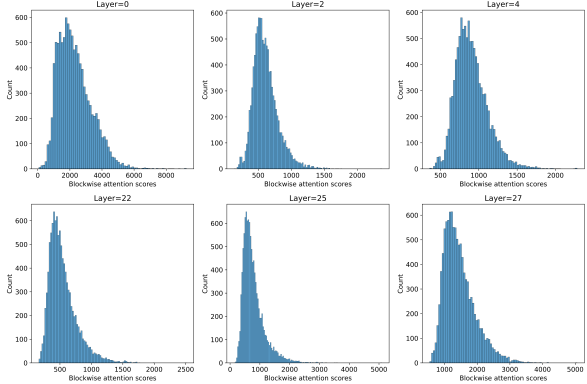}
    \vspace{-3mm}
    \caption{Blockwise attention score distributions across selected layers of the LLaMA-3.2-3B-Instruct model. Each histogram shows the sum of attention scores received by individual blocks (excluding the first block) within a block of length 128 during the prefill phase. The variation in score ranges across layers indicates differing degrees of token mixing, particularly among non-sink tokens.}
    \label{fig:attn_dist}
    \vspace{-3mm}
\end{figure}

\begin{figure}[h]
    \centering
    \includegraphics[width=0.95\linewidth]{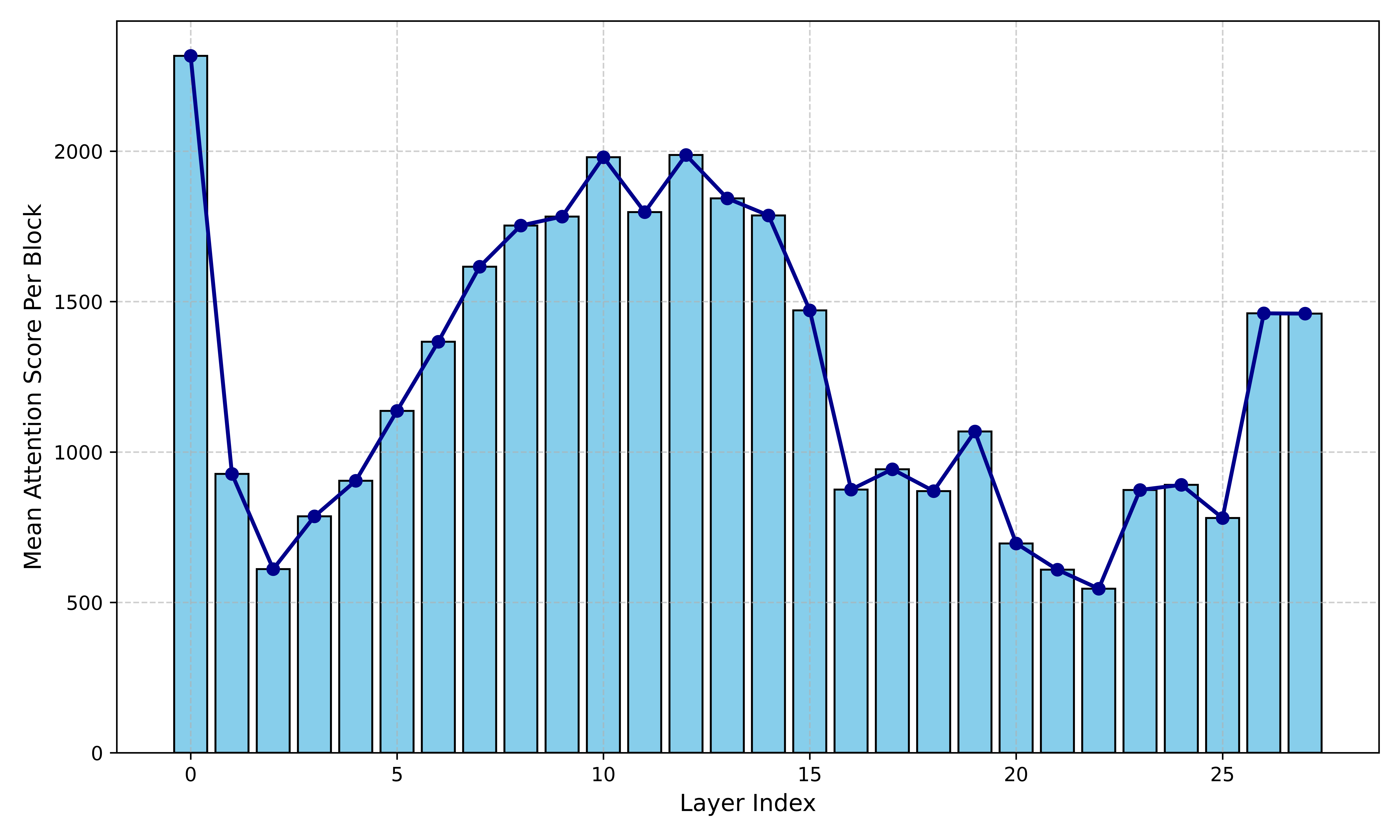}
    \vspace{-3mm}
    \caption{Mean attention scores per block during prefill across layers of the LLaMA-3.2-3B-Instruct model. Each value represents the average attention received by blocks of size 128, excluding the first block. These means serve as a proxy for the extent to which each layer contributes to mixing token embeddings, with higher values indicating stronger integration of non-sink tokens.}
    \vspace{-3mm}
    \label{fig:attn_layer_means}
\end{figure}

To implement this, we construct a calibration dataset by selecting 128 samples from Minipile with context lengths exceeding 12K tokens. For each layer (Figure \ref{fig:attn_dist} \& \ref{fig:attn_layer_means}), we compute the sum of attention scores received by all blocks except the first (which contains sink tokens). A higher cumulative attention score indicates that the layer plays a more transformative role in shaping token representations, and thus should be preserved with lower sparsity.

More specifically, we define a layerwise importance score \( s_i \) to quantify the relative contribution of layer \( i \) to token mixing. 
Let each transformer layer \( i \in \{1, \dots, L\} \) have \( H \) attention heads, and let 
\( A_i^{(h)} \in \mathbb{R}^{N \times N} \) denote the attention matrix for head \( h \) at layer \( i \), 
where \( N \) is the sequence length. 
Let \( \mathcal{B}_1 \) denote the first block (containing sink tokens), 
and \( \mathcal{B}_{\text{non-sink}} \) denote the set of all remaining blocks.

The layerwise importance score \( s_i \) is computed as the total attention mass \emph{received} by non-sink tokens, averaged over all heads and calibration samples:

\begin{equation}
s_i = \frac{1}{|\mathcal{D}| H} 
\sum_{X \in \mathcal{D}} 
\sum_{h=1}^{H} 
\sum_{k \in \mathcal{B}_{\text{non-sink}}} 
\sum_{t=1}^{N} 
A_{i,\,t k}^{(h)}(X)
\end{equation}

where:\vspace{-2mm}
\begin{itemize}
    \item \( \mathcal{D} \) is the calibration dataset (128 samples from Minipile), 
    \item \( A_{i,\,t k}^{(h)}(X) \) is the attention weight from query token \( t \) to key token \( k \) in head \( h \) of layer \( i \) for input \( X \), \vspace{-2mm}
    \item and \( k \in \mathcal{B}_{\text{non-sink}} \) restricts the sum to keys in non-sink blocks (i.e., tokens receiving attention). 
\end{itemize}



where \( N_{\text{non-sink}} = \sum_{X \in \mathcal{D}} |\mathcal{B}_{\text{non-sink}}(X)| \) denotes the total number of non-sink tokens in the calibration set.  
A larger \( s_i \) indicates the layer receives more attention mass on non-sink tokens and therefore should be allocated lower sparsity under the overall budget constraint.

We allocate sparsity levels across layers using a linear schedule (Algorithm \ref{alg:sparsity_schedule}) informed by the attention score sums, $\{s_i\}$ and a predefined sparsity budget. This approach improves performance, particularly in scenarios where the error compensator is less effective. Overall, dynamic sparsity control provides a robust mechanism for balancing efficiency and accuracy, as demonstrated in our ablation studies.

We keep the first block to be always dense because it contains sink tokens and the last block to be dense as well which helps in QA tasks.

\begin{algorithm}[H]
\caption{Layerwise Sparsity Schedule Allocation}
\label{alg:sparsity_schedule}
\begin{algorithmic}[1]
\Require Layerwise importance scores $S = \{s_1, s_2, \dots, s_{L}\}$, overall budget $B$, number of layers $L$
\Ensure Layerwise budgets $\{b_1, b_2, \dots, b_{L}\}$
\State $T \gets B \times L$ \Comment{Initialize total budget}
\State $b \gets [\;]$ \Comment{Empty list for layerwise budgets}
\State $S_{total} \gets \sum_{i=1}^{L} s_i$
\For{$i = 1 \; \textbf{to} \; L$}
    \State $b_i \gets \min\left(1,\; \frac{s_i}{S_{total}} \times T \right)$
    \State $T \gets T - b_i$
    \State $S_{total} \gets S_{total} - s_i$
    \State Append $b_i$ to $b$
\EndFor
\State \Return $b$
\end{algorithmic}
\end{algorithm}

\section{Experiments}
\vspace{-2mm}
We evaluate our approach on English-language tasks from the LongBench benchmark, which is designed to assess model performance in short- to moderate-context settings (4K–32K tokens) and features an average English prompt length of 6,711 words. Since our method specifically targets this range of prompt lengths, LongBench serves as a representative and challenging testbed for evaluation.

Our experiments are conducted using the LLaMA-3 Instruct models \cite{grattafiori2024llama3herdmodels}, specifically the 1B, 3B, and 8B variants (versions 3.2 and 3.1, respectively and Qwen3-4B model \cite{yang2025qwen3technicalreport}. These models offer a diverse range of capacities, allowing us to assess the scalability and generalizability of our sparsity and error compensation techniques across different model sizes and families.

All evaluations focus on downstream task performance, with particular attention to how dynamic sparsity control and error compensation affect accuracy and efficiency under constrained computational budgets. Detailed results and comparisons are presented in Section~\ref{sec:results}, along with ablation studies highlighting the contribution of each component.

\section{Results}\label{sec:results}
We evaluate FFN sparsity across three model sizes on the LongBench benchmark \cite{bai2024longbenchbilingualmultitaskbenchmark}, comparing sparse configurations to their dense counterparts. As shown in Table~\ref{tab:longbench_results_improved}, all models maintain strong performance even at high sparsity levels. At 50\% FFN sparsity, the Llama3.2-1B, Llama3.2-3B, and Llama3.1-8B models incur only a 5.1\%, 4.9\%, and 6.0\% drop in average accuracy, respectively, relative to their dense baselines. Qwen3-4B incurs an even smaller drop of 2.18\% at 50\% sparsity. Smaller sparsity levels (30–40\%) reduce this gap to below 2–4\%. These results indicate that structured FFN sparsity, guided by our expert predictor and error compensator, achieves significant parameter reduction while preserving most of the dense model’s capability across diverse reasoning, summarization, and code tasks.

FFN sparsity also transfers effectively to generation with negligible accuracy loss. Table \ref{tab:longbench_prefill_and_gen_small} indicates that accuracy under both prefill and generation sparsity remains close to the dense baseline. This is enabled by the expert predictor, which generalizes beyond its training on 128-token blocks to arbitrary sequence lengths via its attention-mixing mechanism, and by the error compensator, which corrects token-level errors individually. Consequently, the approach extends to Longbench and MMLU generation tasks without additional training, provided sequence lengths remain short to moderate. 

\begin{figure}[!htpb]
    \centering
    \includegraphics[width=\linewidth]{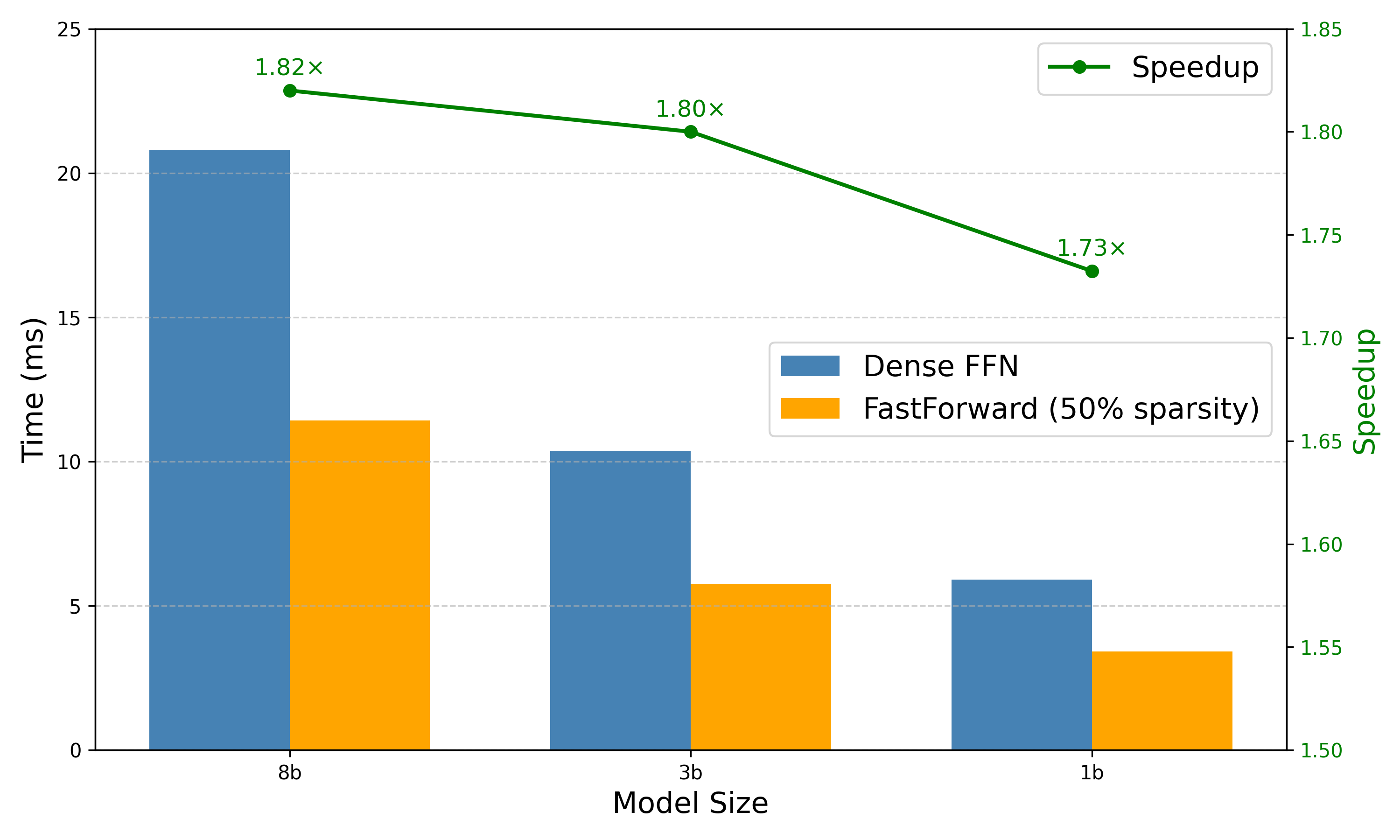}
    \vspace{-3mm}
    \caption{Speedup of the FFN module of Llama3 models with FastForward at 50\% sparsity, profiled on an A5000 GPU using custom CUDA kernels.}
    \label{fig:ffn_kernel}
    \vspace{-5mm}
\end{figure}

\begin{figure}[h]
    \centering
    \vspace{-3mm}
    \includegraphics[width=0.98\linewidth]{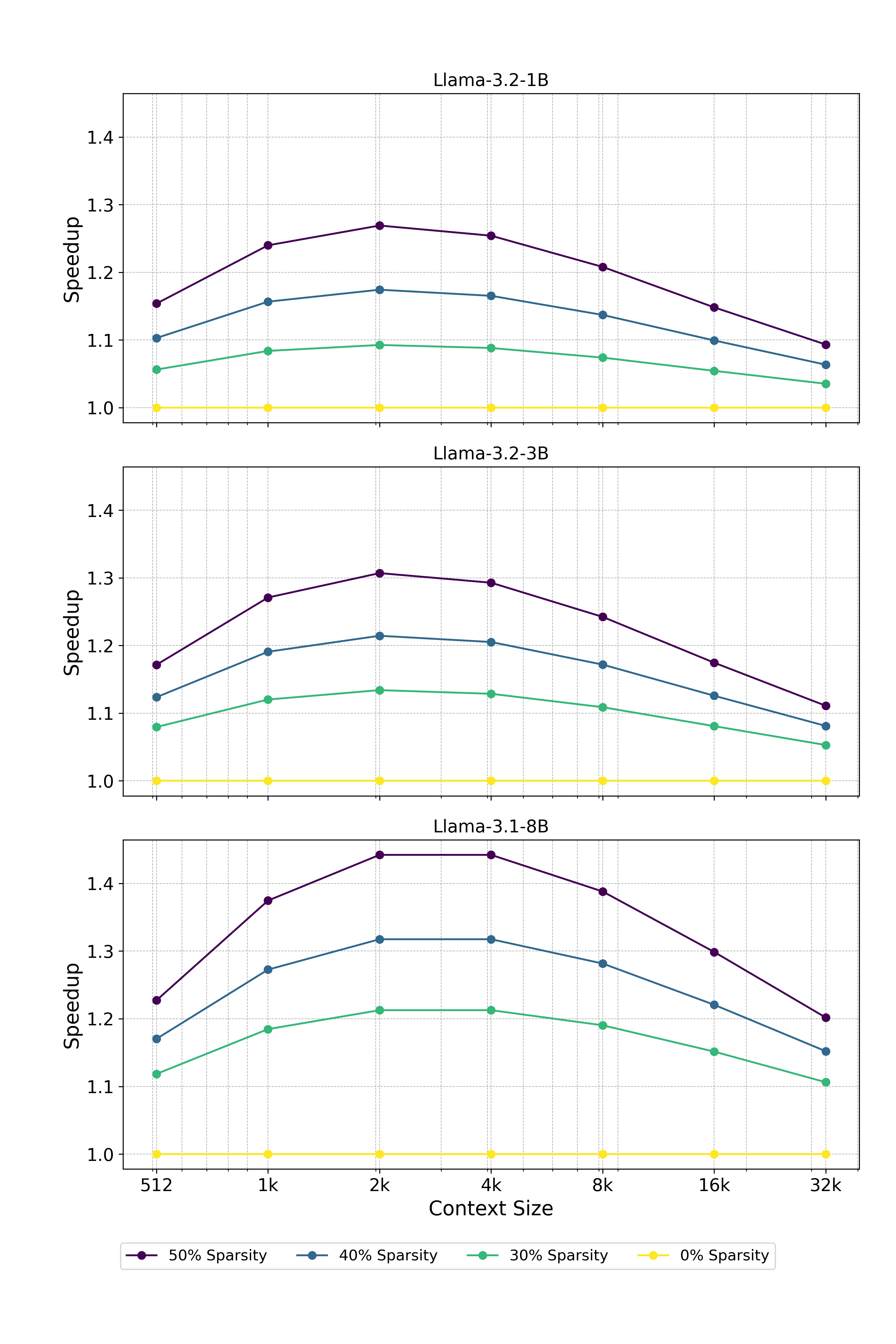}
    \vspace{-5mm}
    \caption{Compute-bound speedup during prefill of Llama 1B, 3B, and 8B models as a function of context size for various sparsity levels. Speedup is measured relative to a dense (0\% sparsity) baseline for each respective model.}
    \vspace{-3mm}
    \label{fig:speedups}
    \vspace{-4mm}
\end{figure}

Figure~\ref{fig:ffn_kernel} presents the speedups achieved by FastForward on the FFN module relative to the dense baseline, while Figure~\ref{fig:speedups} shows the end-to-end compute-bound speedup from sparsity during the prefill stage across several Llama models. We observe a peak end-to-end speedup of up to 1.45$\times$, corresponding to a 45\% reduction in FLOPs at 50\% sparsity.

At shorter context lengths, the speedup is modest because the first and last blocks are computed without sparsity, reducing the effective proportion of tokens benefiting from FFN pruning. As the context length increases, these dense blocks constitute a smaller fraction of the total computation. The speedup reaches its peak for mid-range contexts (approximately 2k–8k tokens), where the linearly scaling FFN layers contribute the most to overall computation. For longer contexts, the speedup gradually diminishes as the quadratic-cost attention operation dominates the total FLOPs, thereby limiting the impact of FFN sparsity.

\begin{table*}[t] 
\centering
\setlength{\tabcolsep}{4pt} 

\sisetup{
    table-format=2.2,      
    table-space-text-post={**} 
}

\begin{tabular}{@{} l l S S S S S S S S[table-format=-1.2] @{}}
\toprule
& & \multicolumn{6}{c}{\textbf{Longbench Task Performance}} & \multicolumn{2}{c}{\textbf{Overall}} \\
\cmidrule(lr){3-8} \cmidrule(lr){9-10}
\textbf{Model} & \textbf{Sparsity Method} & {\makecell{Single\\Doc QA}} & {\makecell{Multi\\Doc QA}} & {\makecell{Summ.}} & {\makecell{Few-shot}} & {\makecell{Synth.}} & {\makecell{Code}} & {\textbf{Average}} & {\makecell{Rel. Gap\\(\%)}} \\
\midrule
Llama3.2-1b & Dense (0\%) & 26.76 & 24.74 & 25.43 & 61.35 & 3.59 & 38.18 & 29.88 & 0.00 \\
 & 30\% & 24.80 & 25.07 & 24.77 & 58.97 & 3.31 & 42.01 & 29.58 & -0.99 \\
 & 40\% & 25.19 & 24.96 & 24.73 & 57.94 & 2.76 & 42.24 & 29.38 & -1.66 \\
 & 50\% & 24.63 & 23.04 & 24.39 & 57.38 & 1.48 & 40.71 & 28.37 & -5.07 \\
\midrule
Llama3.2-3b & Dense (0\%) & 38.65 & 36.72 & 27.67 & 67.85 & 47.29 & 51.59 & 44.72 & 0.00 \\
 & 30\% & 38.25 & 38.63 & 26.85 & 64.93 & 42.04 & 53.57 & 43.74 & -2.18 \\
 & 40\% & 37.17 & 35.95 & 27.45 & 65.44 & 40.34 & 53.82 & 43.03 & -3.76 \\
 & 50\% & 36.39 & 36.37 & 26.53 & 65.31 & 36.67 & 56.37 & 42.54 & -4.86 \\
\midrule
Llama3.1-8b & Dense (0\%) & 43.25 & 45.43 & 29.19 & 69.44 & 54.13 & 59.07 & 49.76 & 0.00 \\
 &  30\% & 42.97 & 40.80 & 28.73 & 67.36 & 52.59 & 59.10 & 48.22 & -3.09 \\
 &  40\% & 43.86 & 39.09 & 28.47 & 67.09 & 51.64 & 56.08 & 47.40 & -4.75 \\
 & 50\% & 42.32 & 38.16 & 28.21 & 66.31 & 51.39 & 56.29 & 46.78 & -5.99 \\
\midrule
Qwen3-4b & Dense (0\%) & 29.21 & 17.24 & 25.39 & 69.33 & 50.96 & 59.16 & 41.26 & 0.00 \\
 & 30\% & 28.45 & 16.48 & 25.24 & 69.30 & 50.75 & 59.10 & 40.92 & -0.82 \\
 & 40\% & 28.32 & 15.71 & 25.23 & 68.99 & 50.81 & 59.00 & 40.71 & -1.33 \\
 & 50\% & 27.45 & 15.64 & 25.03 & 68.74 & 50.79 & 58.26 & 40.36 & -2.18 \\
\bottomrule
\end{tabular}
\caption{
    \textbf{Performance comparison of different FFN sparsity configurations across Longbench tasks and model sizes.} All sparse models in this study utilize an expert predictor, an error corrector, keep the first and last blocks dense and apply a layerwise sparsity schedule according to Algorithm \ref{alg:sparsity_schedule}. We compare a dense baseline (0\%) against various sparsity levels. The 'Rel. Gap (\%)' column shows the performance change relative to the dense baseline for each model size. Abbreviations: Summarization (Summ.), Synthetic (Synth.).
}
\label{tab:longbench_results_improved}
\vspace{-3mm}
\end{table*}

\begin{table}[!htpb]
\centering
\small 
\setlength{\tabcolsep}{3pt} 
\captionsetup{skip=2pt} 
\sisetup{table-format=2.2}
\begin{tabular}{@{} l l S S @{}}
\toprule
\textbf{Model} & \textbf{Sparsity Method} & {\makecell{\textbf{Longbench} \\ \textbf{(Avg)}}} & {\makecell{\textbf{MMLU} \\ \textbf{(Avg)}}} \\
\midrule
1b & Dense (0\%) & 29.88 & 46.06 \\
1b & Sparse (50\%) & 28.31 & 45.35 \\
\midrule
3b & Dense (0\%) & 44.72 & 60.68 \\
3b & Sparse (50\%) & 42.46 & 60.16 \\
\midrule
8b & Dense (0\%) & 49.76 & 67.84 \\
8b & Sparse (50\%) & 46.92 & 67.17 \\
\bottomrule
\end{tabular}
\caption{
    \textbf{Performance of Sparse Llama3 Models in Both Prefill and Generation.}
    This table compares dense models (0\% sparsity) against our sparse models (50\%) across three benchmarks. The sparse method applies FFN sparsity during both the prefill and generation phases, utilizing a single expert predictor and error compensator for both, combined with a layer-wise sparsity schedule.
}
\label{tab:longbench_prefill_and_gen_small}
\vspace{-3mm}
\end{table}


\section{Ablation}
\vspace{-2mm}
This section presents an ablation study of the components of \textsc{FastForward} to identify the factors contributing to its effectiveness.
\vspace{-2mm}
\subsection{Layerwise Sparsity Schedule}
Table \ref{tab:layerwise_ablation} shows that assigning layer-specific sparsity budgets based on their roles in token mixing, as described in Algorithm~\ref{alg:sparsity_schedule}, improves performance by 1–3\% compared to applying a uniform sparsity level across all layers. This improvement stems from the fact that certain layers contribute more significantly to mixing non-sink tokens, and preserving these layers with higher fidelity enhances downstream performance.
\vspace{-2mm}
\begin{table}[!htbp]
\centering
\captionsetup{skip=2pt}
\begin{tabular}{|c|l|c|}
\hline
\textbf{Model} & \textbf{Sparsity} & \textbf{Average} \\
\hline
Llama3.2-1b & Layerwise 50\% & 28.37 \\
 & Uniform 50\% & 28.08 \\
\hline
Llama3.2-3b & Layerwise 50\% & 42.54 \\
 & Uniform 50\% & 41.33 \\
\hline
Llama3.1-8b & Layerwise 50\% & 46.78 \\
 & Uniform 50\% & 46.42 \\
\hline
Qwen3-4b & Layerwise 50\% & 40.36 \\
 & Uniform 50\% & 40.00 \\
\hline
\end{tabular}
\caption{An ablation study on the effect of layerwise sparsity schedule for FastForward during prefill.}
\label{tab:layerwise_ablation}
\vspace{-3mm}
\end{table}

\subsection{First and Last Block Dense}
Table \ref{tab:first_last_dense_table} shows that keeping the first and last blocks dense improves performance by 15–30\% relative to the sparse baseline. The improvement arises because these blocks contain tokens that are critical for attention and prompt encoding, making them less tolerant to sparsification.

\begin{table}[h!]
\centering
\begin{tabular}{|l|r|}
\hline
 \textbf{Sparsity Configuration} & \textbf{Average} \\
\hline
\multicolumn{2}{|c|}{Llama-1B} \\
\hline
  Uniform (50\%) & 20.41 \\
   + w/ Dense First & 23.99 \\
   + w/ Dense First \& Last & 28.08 \\
\hline
\multicolumn{2}{|c|}{Llama-3B} \\
\hline
  Uniform (50\%) all blocks & 37.72 \\
   + w/ Dense First  & 37.40 \\
   + w/ Dense First \& Last & 41.33 \\
\hline
\multicolumn{2}{|c|}{Llama-8B} \\
\hline
  Uniform (50\%) & 42.37 \\
   + w/ Dense First & 43.43 \\
   + w/ Dense First \& Last & 46.42 \\
\hline
\end{tabular}
\caption{An ablation study on the effect of keeping certain blocks dense during prefill. "Uniform" applies 50\% sparsity everywhere, while "+ w/ Dense Last" and "+ w/ Dense First \& Last" keep the specified blocks dense. The metric shown is the average performance (lower is better).}
\label{tab:first_last_dense_table}
\vspace{-5mm}
\end{table}

\subsection{Error compensator}
\vspace{-2mm}
Table \ref{tab:model_sparsity_errorcomp} reports results with and without the error compensator. The error compensator consistently improves performance across all models. Notably, the estimated error vectors tend to have very small norms, reflecting the model’s limited certainty about the exact error direction. In practice, it applies corrections only along the most confident direction, resulting in weak but consistent adjustments. This property enables the error compensator to remain effective across different sparsity levels, which is a practical advantage.
\vspace{-3mm}
\subsection{Expert Predictor}
\vspace{-2mm}
Table \ref{tab:expert_ablation_concise} evaluates the effect of the expert predictor through three variants: 
(i) a \textbf{first-block static predictor}, which applies the GRIFFIN method to select experts for all subsequent blocks using statistics from the first block; 
(ii) a \textbf{per-block dynamic predictor}, which selects experts for each block based on its own prompt statistics—an \textit{oracle} approach that is infeasible in practice since it requires computing dense FFN outputs for all blocks; and 
(iii) a \textbf{trained predictor}, which learns to approximate the per-block dynamic’s expert selection behavior.

\begin{table}[h!]
\centering
\begin{tabular}{|c|l|c|}
\hline
\textbf{Model} & \textbf{Sparsity} & \textbf{Average} \\
\hline
1b & 50\% & 28.08 \\
1b & 50\% - error compensator & 27.50 \\
\hline
3b & 50\% & 41.33 \\
3b & 50\% - error compensator & 38.13 \\
\hline
8b & 50\% & 46.42 \\
8b & 50\% - error compensator & 46.01 \\
\hline
\end{tabular}
\caption{An ablation study on the effect of error compensator module for FastForward during prefill on Llama3 models.}
\label{tab:model_sparsity_errorcomp}
\vspace{-3mm}
\end{table}

Results show that the \textbf{trained predictor} achieves performance very close to the per-block dynamic while substantially outperforming the GRIFFIN-based first-block predictor. In particular, for the 3B model, the trained predictor even exceeds the dynamic per-block, as it learns to identify expert neurons from aggregated cross-prompt statistics rather than relying on the local statistics of a single prompt.

\begin{table}[h!]
\centering
\begin{tabular}{@{} |l| l| c| @{}}
\toprule
\textbf{Model} & \textbf{Predictor / Sparsity} & \textbf{Average} \\
\midrule
1b & 50\% (Trained Predictor) & 23.99 \\
1b & 50\% (Per-Block Dynamic) & 26.02 \\
1b & 50\% (First-Block Static) & 18.13 \\
\midrule
3b & 50\% (Trained Predictor) & 37.40 \\
3b & 50\% (Per-Block Dynamic) & 36.45 \\
3b & 50\% (First-Block Static) & 25.06 \\
\midrule
8b & 50\% (Trained Predictor) & 43.43 \\
8b & 50\% (Per-Block Dynamic) & 44.75 \\
8b & 50\% (First-Block Static) & 30.07 \\
\bottomrule
\end{tabular}
\caption{
    \label{tab:expert_ablation_concise}
    \textbf{Ablation study of expert prediction methods.}
    We compare three predictors on Llama3 models, all using a dense FFN for the first block and 50\% sparsity for all subsequent blocks. The \textbf{Per-Block Dynamic} predictor acts as an upper bound, using per-block GRIFFIN statistics \cite{dong2024promptpromptedadaptivestructuredpruning} for ideal expert selection. The \textbf{First-Block Static} predictor is a baseline that reuses the expert selection from the first block for all subsequent blocks. Our \textbf{Trained Predictor} is a model trained to emulate the Per-Block Dynamic's selections (Section~\ref{sec:expert_pred}).
}
\vspace{-3mm}
\end{table} 

%

\section{Discussion and Conclusion}
Recent efforts in model efficiency have predominantly focused on sparsifying FFN layers during the auto-regressive generation phase. In this work, we shift the focus to the prefill stage, demonstrating that significant computational savings can be achieved by applying structured FFN sparsity during block prompt processing. This is a critical area for optimization, as FFN computations account for more than 50\% of the total forward pass FLOPs for the small and moderate context lengths typical of prefill. While the computational bottleneck eventually shifts to the attention mechanism at very large sequence lengths, FFNs remain a significant and optimizable component.

The central principle of our method is exploiting the structured activation patterns within FFNs—a phenomenon termed neuron "flocking," \cite{dong2024promptpromptedadaptivestructuredpruning} where only a coherent subset of neurons show high activation for a given context. Our work is novel in its approach to operationalize this insight for prefill. We introduce a system that integrates three key technical contributions to achieve high-fidelity sparse computation. 

First, an \textit{expert predictor} proactively identifies the most influential neurons for a given input context before the main computation begins. Second, to preserve model accuracy, an \textit{error compensator} corrects for consistent biases introduced by the sparsification process. Finally, a \textit{layerwise sparsity schedule} intelligently distributes the fixed compute budget across the model, allocating more resources to layers that are more critical for the model's transformations, thereby maximizing performance for a given level of sparsity.
\vspace{-2mm}

\section{Limitations and Future Work}

Our work presents a promising approach to FFN sparsity, but there are clear limitations that offer avenues for future research.

A primary limitation lies in the current design of the error compensator. The compensator produces corrections of a small norm and, critically, operates without any information regarding which expert neurons were selected or pruned for a given block. This lack of a direct signal inhibits its ability to counteract significant errors that may arise from suboptimal expert selection. Future work should focus on enriching the compensator with this information, potentially by conditioning its output on the expert selection mask. This could enable it to make more targeted and substantial corrections, thereby improving the model's overall robustness to sparsity.

Enhancing the error compensator could also unlock significant efficiency gains by enabling a shift from dynamic to static experts. Our current model dynamically loads different expert weights for each block, which is intensive on memory bandwidth. A more powerful error compensator, capable of correcting larger deviations, could allow the use of a single, static set of experts—identified from the first prompt block—for all subsequent computations. This would drastically reduce memory bandwidth requirements during inference, as the same expert weights would be used throughout an entire sequence.


\bibliography{example_paper}
\bibliographystyle{mlsys2025}

\end{document}